%% file: acl2018.tex
\documentclass[11pt,a4paper]{article}
\usepackage[hyperref]{acl2018}
\usepackage{times}
\usepackage{latexsym}

\usepackage{url}

\usepackage{algorithm}
\usepackage{algorithmic}
\usepackage{amssymb}
\usepackage{amsmath}

\usepackage{multirow}
\usepackage{multicol}

\usepackage[pdftex]{graphicx} 
\graphicspath{{figures/}}

\usepackage{amsfonts}       % blackboard math symbols
\usepackage{nicefrac}       % compact symbols for 1/2, etc.
\usepackage{microtype}      % microtypography

\usepackage{float}
\aclfinalcopy % Uncomment this line for the final submission
\def\aclpaperid{870} %  Enter the acl Paper ID here

\title{Fighting Offensive Language on Social Media \\ with Unsupervised Text Style Transfer}

\author{Cicero Nogueira dos Santos\thanks{\ \ Equal contribution.} \\
  IBM Research \\
  T.J. Watson Research Center \\
  {\tt cicerons@us.ibm.bom} \\\And
  Igor Melnyk\footnotemark[1] \\
  IBM Research \\
  T.J. Watson Research Center \\
  {\tt igor.melnyk@ibm.com} \\\And
  Inkit Padhi\footnotemark[1] \\
  IBM Watson \\
  T.J. Watson Research Center \\
  {\tt inkit.padhi@ibm.com} \\}

\date{}

\begin{document}
\maketitle
\begin{abstract}
We introduce a new approach to tackle the problem of offensive language in online social media. Our approach uses unsupervised text style transfer to \emph{translate} offensive sentences into non-offensive ones. We propose a new method for training encoder-decoders using non-parallel data that combines a collaborative classifier, attention and the cycle consistency loss. Experimental results on data from Twitter and Reddit show that our method outperforms a state-of-the-art text style transfer system in two out of three quantitative metrics and produces reliable non-offensive transferred sentences.
\end{abstract}

\input{introduction}
\input{method}

\input{related_work}
\input{experiments}

\input{conclusions}

%\section*{Acknowledgments}

% include your own bib file like this:
%\bibliographystyle{acl}
%\bibliography{acl2018}
\bibliography{acl2018}
\bibliographystyle{acl_natbib}

%\appendix

% \section{Supplemental Material}
% \label{sec:supplemental}
% ACL 2018 also encourages the submission of supplementary material

\end{document}

%% file: introduction.tex
\section{Introduction}
The use of offensive language is a common problem of abusive behavior on online social media networks.
Various work in the past have attacked this problem by using different machine learning models to detect abusive behavior  \cite{xiang:2012,warner:2012,kwok:2013,wang:2014,nobata:2016,burnap2015,davidson2017,founta2018}.
Most of these work follow the assumption that it is enough to filter out the entire offensive post.
%However, 
%a user that is consuming some online content may not want to filter out the message entirely but instead have it in a style that is non-offensive and still be able to read it. 
However,
a user that is consuming some online content may not want an entirely filtered out message but instead have it in a style that is non-offensive and still be able to comprehend it in a polite tone.
%On the other hand,
%some users would think twice before posting something rude on the Internet if one could not only alert him/her that a content is offensive, but also offer a non-offensive version of the post.
On the other hand, for those users who plan to post an offensive message, if one could not only alert that a content is offensive and will be blocked, but also offer a polite version of the message that can be posted, this could encourage many users to change their mind and avoid the profanity.
%

%Among different the types of abusive behaviors, the use of offensive language is one that can affect both the user that is producing the content as well as the user consuming the content. 

%Translating offensive posts into non-offensive ones can be useful in different situations: (a); (b); 

In this work we introduce a new way to deal with the problem of offensive language on social media. Our approach consists on using style transfer techniques to \emph{translate} offensive sentences into non-offensive ones.
A simple encoder-decoder with attention \cite{bahdanau14} would be enough to create a reasonable translator if a large parallel corpus is available. However, unlike machine translation, to the best of our knowledge, there exists no dataset of parallel data available for the case of offensive to non-offensive language. Moreover, it is important that the transferred text uses a vocabulary that is common in a particular application domain. Therefore, unsupervised methods that do not use parallel data are needed to perform this task.

We propose a method to perform text style transfer addressing two main challenges arising when using non-parallel data in the encoder-decoder framework: (a) there is no straightforward way to train the encoder-decoder because we cannot use maximum likelihood estimation on the transferred text due to lack of ground truth; (b) it is difficult to preserve content while transferring the input to a new style. 
We address (a) using a single collaborative classifier, as an alternative to commonly used adversarial discriminators, e.g., as in \cite{shen17}.
We approach (b) by using the attention mechanism combined with a cycle consistency loss. 

In this work we also introduce two benchmark datasets for the task of transferring offensive to non-offensive text that are based on data from two popular social media networks: Twitter and Reddit. We compare our method to the approach of \citet{shen17} using three quantitative metrics: classification accuracy, content preservation and perplexity. Additionally, some qualitative results are also presented with a brief error analysis.

%% file: method.tex
\section{Method}
\begin{figure*}[!t]
	\centering
	\includegraphics[width=1.0\textwidth]{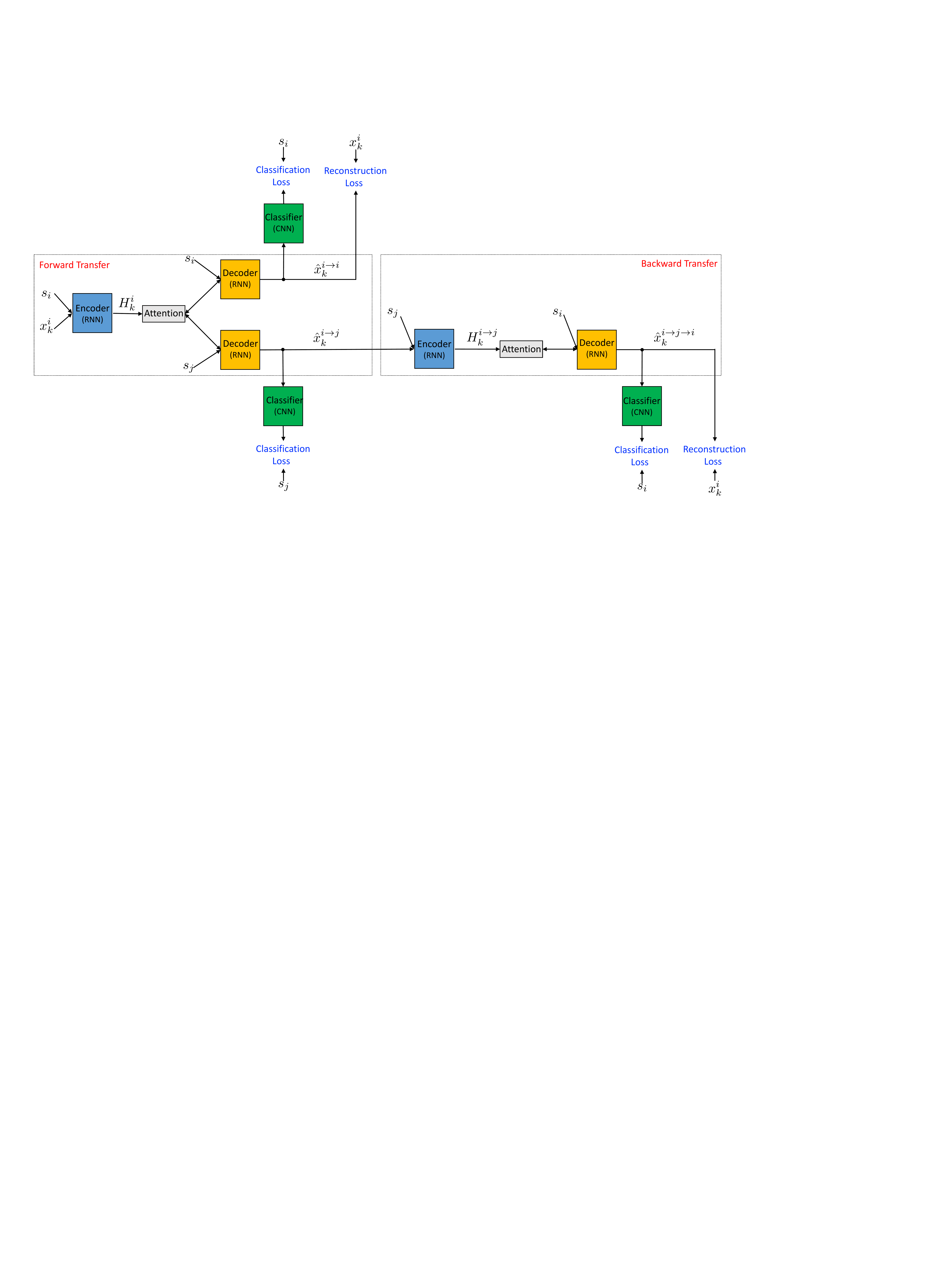}
\vspace{-1em}        
	\caption{Proposed framework of a Neural Text Style Transfer algorithm using non-parallel data.}
	\label{fig:model_training}
\end{figure*}

We assume access to a text dataset consisting of two non-parallel corpora $X = X_0 \cup X_1$ with different style values $s_0$ and $s_1$ (offensive and non-offensive) of a total of $N=m+n$ sentences, where $|X_0| = m$ and $|X_1|=n$. We denote a randomly sampled sentence $k$ of style $s_i$ from $X$ as $x_k^{i}$, for $k \in 1, \ldots, N$ and $i \in \{0,1\}$. 
A natural approach to perform text style transfer is to use a regular encoder-decoder network. However, the training of such network would require parallel data. Since in this work we consider a problem of unsupervised style transfer on non-parallel data, we propose to extend the basic encoder-decoder by introducing a collaborative classifier and a set of specialized loss functions that enable the training on such data. Figure \ref{fig:model_training} shows an overview of the proposed style transfer approach. Note that for clarity, in Figure \ref{fig:model_training} we have used multiple boxes to show encoder, decoder and classifier, the actual model contains a single encoder and decoder, and one classifier.

As can be seen from Figure \ref{fig:model_training}, the encoder (a GRU RNN, $E(x_k^{i}, s_i)=H_k^i$) takes as input a sentence $x_k^{i}$ together with its style label $s_i$, and outputs $H_k^i$, a sequence of hidden states. The decoder/generator (also a GRU RNN, $G(H_k^i, s_j) = \hat{x}_k^{i\rightarrow j}$ for $i,j \in {0,1}$) takes as input the previously computed $H_k^i$ and a desired style label $s_j$ and outputs a sentence $\hat{x}_k^{i\rightarrow j}$,  which is the original sentence but transferred from style $s_i$ to style $s_j$. The hidden states $H_k^i$ are used by the decoder in the attention mechanism \cite{bahdanau14}, and in general can improve the quality of the decoded sentence. When $i=j$, the decoded sentence $\hat{x}_k^{i\rightarrow i}$ is in its original style $s_i$ (top part of Figure \ref{fig:model_training}); for $i \neq j$, the decoded/transferred sentence $\hat{x}_k^{i\rightarrow j}$ is in a different style $s_j$ (bottom part of Figure\ref{fig:model_training}). Denote all transferred sentences as $\hat{X} = \{\hat{x}_k^{i\rightarrow j} ~|~ i\neq j, k=1,\ldots, N\}$. The classifier (a CNN), then takes as input the decoded sentences and outputs a probability distribution over the style labels, i.e., $C(\hat{x}_k^{i\rightarrow j}) = p_C(s_j|\hat{x}_k^{i\rightarrow j})$ (see Eq. \eqref{eq:class_td}). By using the collaborative classifier our goal is to produce a training signal that indicates the effectiveness of the current decoder on transferring a sentence to a given style.

Note that the top branch of Figure \ref{fig:model_training} can be considered as an auto-encoder and therefore we can enforce the closeness between $\hat{x}_k^{i\rightarrow i}$ and $x_k^i$ by using a standard cross-entropy loss (see Eq. \eqref{eq:rec_loss}). However, for the bottom branch, once we transferred $X$ to $\hat{X}$ (forward-transfer step), due to the lack of parallel data, we cannot use the same approach. For this purpose, we propose to transfer $\hat{X}$ back to $X$ (back-transfer step) and compute the reconstruction loss between $\hat{x}_k^{i\rightarrow j \rightarrow i}$ and $x_k^i$ (see Eq. \eqref{eq:back_rec_loss}. Note also that as we transfer the text forward and backward, we also control the accuracy of style transfer using the classifier (see Eqs. \eqref{eq:class_td}, \eqref{eq:class_od} and  \eqref{eq:class_btd}). In what follows, we present the details of the loss functions employed in training.

\subsection{Forward Transfer}
\textbf{Reconstruction Loss.} 
Given the encoded input sentence $x_k^{i}$ and the decoded sentence $\hat{x}_k^{i\rightarrow i}$, the reconstruction loss measures how well the decoder $G$ is able to reconstruct it:
\begin{align}
\label{eq:rec_loss}
\mathcal L_{rec} = \mathbb{E}_{x_k^i\sim X} \left[\text{-}\log p_G(x_k^i| E(x_k^i, s_i), s_i)\right].
\end{align}

\noindent \textbf{Classification Loss.} Formulated as follows:
\begin{align}
\label{eq:class_td}
\mathcal L_{class\_td} &= \mathbb{E}_{\hat{x}_k^{i \rightarrow j}\sim \hat{X}} \left[-\log p_C(s_j|\hat{x}_k^{i\rightarrow j})\right].
\end{align}
For the encoder-decoder this loss gives a feedback on the current generator's effectiveness on transferring sentences to a new style.
For the classifier, it provides an additional training signal from generated data, enabling the classifier to be trained in a semi-supervised regime.

\noindent \textbf{Classification Loss - Original Data.} In order to enforce a high classification accuracy, the classifier also uses a supervised classification loss, measuring the classifier predictions on the original (supervised) instances $x_{k}^{i} \in X$:
\begin{align}
\label{eq:class_od}
\mathcal L_{class\_od} = \mathbb{E}_{x_k^i\sim X} \left[- \log p_C(s_i|x_k^i)\right].
\end{align}

\subsection{Backward Transfer}
\textbf{Reconstruction Loss.} The \emph{back-transfer (or cycle consistency) loss} \cite{Zhu2017:CycleGAN} is motivated by the difficulty of imposing constraints on the transferred sentences. Back-transfer transforms the transferred sentences $\hat{x}_k^{i\rightarrow j}$ back to the original style $s_i$, i.e., $\hat{x}_k^{i\rightarrow j \rightarrow i}$ and compares them to $x_k^i$. This also implicitly imposes the constraints on the generated sentences and improves the content preservation. The loss is formulated as follows:
\begin{align}
\label{eq:back_rec_loss}
\resizebox{0.42\textwidth}{!}{%
$\mathcal L_{back\_rec} \hspace{-1pt}=\hspace{-1pt} \mathbb{E}_{x_k^i\sim X} \hspace{-2pt}\left[-\log p_G(x_k^i| E(\hat{x}_k^{i\rightarrow j}\hspace{-1pt}, s_j), s_i)\right]$,%
}
\end{align}
\noindent which can be thought to be similar to an auto-encoder loss in \eqref{eq:rec_loss} but in the style domain.

\textbf{Classification Loss.}
Finally, we ensure that the back-transferred sentences $\hat{x}_k^{i\rightarrow j \rightarrow i}$ have the correct style label $s_i$:
\begin{align}
\label{eq:class_btd}
\resizebox{0.42\textwidth}{!}{%
$\mathcal L_{class\_btd} \hspace{-2pt}= \hspace{-2pt}\mathbb{E}_{\hat{x}_k^{i \rightarrow j}\sim \hat{X}} \hspace{-2pt}\left[-\log p_C(s_i|G(E(\hat{x}_k^{i\rightarrow j}, s_j),s_i))\right]$.
}
\end{align}

In summary, the training of the components of our architecture consists in optimizing the following loss function using SGD with back-propagation:
\vspace{-1.5em}
\begin{equation}
\label{eq:total_loss}
\begin{split}
\mathcal L(\theta_E,\theta_G,\theta_{C}) \nonumber 
=\min_{E, G, C} ~ \mathcal L_{rec} + \mathcal L_{back\_rec} \\
\nonumber + \mathcal L_{class\_od}
+ \mathcal L_{class\_td} + \mathcal L_{class\_btd}
\end{split}
\end{equation}

%% file: related_work.tex
\section{Related Work}
Most previous work that address the problem of offensive language on social media has focused on text classification using different machine learning methods \cite{xiang:2012,warner:2012,kwok:2013,wang:2014,burnap2015,nobata:2016,davidson2017,founta2018}.
To the best of our knowledge, there is no previous work on approaching the offensive language problem using style transfer methods.

Different strategies for training encoder-decoders using non-parallel data have been proposed recently.
Many of these methods borrow the idea of using an adversarial discriminator/classifier from the Generative Adversarial Networks (GANs) framework \cite{goodfellow2014GANs} and/or use a cycle consistency loss.
\citet{Zhu2017:CycleGAN} proposed the pioneering use of the cycle consistency loss in GANs to perform image style transfer from non-parallel data. 
In the NLP area, 
some recent effort has been done on the use of non-parallel data for style/content transfer \cite{shen17,melnyk2017,fu:aaai18} and machine translation \cite{lample2018NMT,artetxe2018NMT}.
\citet{shen17},  \citet{fu:aaai18} and \citet{lample2018NMT} use adversarial classifiers as a way to force the decoder to transfer the encoded source sentence to a different  style/language.
\citet{lample2018NMT} and \citet{artetxe2018NMT} use the cycle consistency loss to enforce content preservation in the translated sentences.
Our work differs from the previous mentioned work in different aspects: we propose a new relevant style transfer task that has not been previously explored; our proposed method combines a collaborative classifier with the cycle consistency loss, 
which gives more stable results.
Note that a potential extension to a problem of multiple attributes transfer would still use a single classifier, while in \cite{shen17,fu:aaai18} this may require as many discriminators as the number of attributes.

Another line of research connected to this work consists in the automatic text generation conditioned on stylistic attributes.
\cite{hu2017} and \cite{ficler:2017} are examples of this line of work which use labeled data during training.

% Another line of research connected to this work consists in the automatic text generation conditioned on stylistic attributes.
% \citet{hu2017} combines variational auto-encoders (VAEs) and attribute discriminators to generate text whose attributes are controlled.
% \citet{ficler:2017}
% propose the use of conditional LSTM language models where the conditioning context consists in the desired content and stylistic parameters.
% Both approaches use labeled data during training.

%% file: experiments.tex
\section{Experiments}
\subsection{Datasets}

We created datasets of offensive and non-offensive texts by leveraging \citet{henderson2018}'s preprocessing of Twitter \cite{ritter2010} and Reddit Politics \cite{Iulian2017:DRLChatBot} corpora, which contain a large number of social media posts. 
\citet{henderson2018} have used Twitter and Reddit
datasets to evaluate the impact of offensive language
and hate speech in neural dialogue systems.

We classified each entry in the two datasets using the offensive language and hate speech classifier from \citep{davidson2017}. For Reddit, 
since the posts are long, 
we performed the classification at the sentence level.
We note that since ground truth (parallel data) is not available, 
it is important to use the same classifier for data generation and evaluation so as to have a fair comparison and avoid inconsistencies.
Therefore, 
we use the classifier from \citep{davidson2017} to test the performance of the compared algorithms in Sec. \ref{sec:results}. 

For our experiments, we used sentences/tweets with size between 2 and 15 words and removed repeated entries, which were frequent in Reddit. The final datasets have the following number of instances:
Twitter - train [58,642 / 1,962,224] (offensive / non-ofensive), dev [7842] (offensive), test [7734]; 
Reddit - [224,319 / 7,096,473], dev [11,883], test [30,583] .
In both training sets the number of non-offensive entries is much larger than of the offensive ones, which is not a problem since the objective is to have the best possible transfer to the non-offensive domain.
We limited the vocabulary size by using words with frequency equal or larger than 70 (20) in Reddit (Twitter) dataset.
All the other words are replaced by a placeholder token.

\subsection{Experimental Setup}

In all the presented experiments, we have used the same model parameters and the same configuration: 
the encoder/decoder is a single layer GRU RNN with 200 hidden neurons;
the classifier is a single layer CNN with a set of filters of width 1, 2, 3 and 4, and size 128 (the same configuration as in the discriminators of \cite{shen17}).
Following \cite{shen17}, we have also used randomly initialized word embeddings of size 100, and trained the model using Adam optimizer with the minibatch size of 64 and learning rate of 0.0005.
The validation set has been used to select the best model by early stopping. Our model has a quite fast convergence rate and achieves good results within just 1 epoch for the Reddit dataset and 5 epochs for the Twitter dataset.

Our baseline is the model of \citet{shen17}\footnote{https://github.com/shentianxiao/language-style-transfer} and it has been used with the default hyperparameter setting proposed by the authors. 
We have trained the baseline neural net for three days using a K40 GPU machine, 
corresponding to about 13 epochs on the Twitter dataset and 5 epochs on the Reddit dataset.
The validation set has also been used to select the best model by early stopping.

%Unlike ours, this model took much longer to train (three days on K40 GPU, corresponding to about 5 epochs on the Reddit dataset and 13 epochs on the Twitter dataset). The validation set has also been used to select the best model by early stopping.

\subsection{Results and Discussion}
\label{sec:results}

Although the method proposed in this paper can be used to transfer text in both directions,
we are interested in transferring in the direction of offensive to non-offensive only. Therefore, all the results reported in this section correspond to this direction.

In Table \ref{table:acc_ppl},
we compare our method with the approach of \citet{shen17} using three quantitative metrics:
(1) \emph{classification accuracy} (Acc.),
which we compute by applying \citet{davidson2017}'s classifier to the transferred test sentences;
(2) \emph{content preservation} (CP),
a metric recently proposed by \citet{fu:aaai18} which uses pre-trained word embeddings to compute the content similarity between transferred and original sentences.
We use Glove embeddings of size 300 \cite{pennington2014glove};
(3) \emph{perplexity} (PPL),
which is computed by a word-level LSTM language model trained using the non-offensive training sentences.
%\footnote{https://github.com/pytorch/examples/tree/master/word\_language\_model}

\begin{table}[!ht]
\begin{center}
\begin{tabular}{|l|l|ccc|}
\hline \bf Dataset & \bf System & \bf Acc. & \bf CP  & \bf PPL \\ \hline
\multirow{2}{*}{Reddit} & [Shen17] & 87.66 & 0.894 & \bf 93.59 \\
                        & Ours & \bf 99.54 & \bf 0.933 & 115.75 \\
\hline
\multirow{2}{*}{Twitter} & [Shen17] & 95.36 & 0.891 & \bf 90.97 \\
                         & Ours & \bf 99.63 & \bf 0.947 & 162.75 \\
\hline
\end{tabular}
\end{center}
\caption{Classification accuracy, content preservation and perplexity for two datasets.}
\label{table:acc_ppl}
\end{table}

\begin{table*}[ht!]
\small
\begin{center}
\begin{tabular}{|l|l|l|}
\hline \bf  & \multicolumn{1}{|c|}{ \bf Reddit} & \multicolumn{1}{|c|}{\bf Twitter} \\
\hline \bf Original & \it for f**k sake , first world problems are the worst & i 'm back bitc**s ! ! !  \\ \hline
\cite{shen17} & for the money , are one different countries & i 'm back ! ! ! \\
Ours & for hell sake , first world problems are the worst & i 'm back bruh ! ! ! \\
\hline \bf Original & \it what a f**king circus this is . & \it lol damn imy fake as* lol \\ \hline
\cite{shen17} & what a this sub is bipartisan . & lol damn imy sis lol \\
Ours        & what a big circus this is . & lol dude imy fake face lol \\
\hline \bf Original & \it i hope they pay out the as* , fraudulent or no . & \it bros before hoes \\ \hline
\cite{shen17} & i hope the work , we out the UNK and no . & club tomorrow \\
Ours        & i hope they pay out the state , fraudulent or no . & bros before money \\
\hline
\end{tabular}
\end{center}
\caption{Example of offensive sentences from Reddit and Twitter and their respective transferred versions.}
\label{table:transf_examp}
\end{table*}

\begin{table}[!ht]
\begin{center}
\begin{tabular}{|l|}
\hline 
%\bf  \bf Sentence \\ \hline
%\it what f***ing century are you living in ? \\
what \emph{big} century are you living in ? \\
life is so \emph{big} cheap to some people .\\
you 're \emph{big} pathetic . \\
\hline
% \it can we just nuke these f**ks already ? \\
% can we just nuke these cops already ? \\
% \hline
% \it pope is bad a** . \\
% pope is bad state . \\
% \hline
\end{tabular}
\end{center}
\caption{Examples of common mistakes made by our proposed model.}
\label{table:mistakes}
\end{table}

As can be seen from the table, our proposed method achieves high accuracy on both datasets,
which means that almost 100\% of the time \citet{davidson2017}'s classifier detects that the transferred sentences are non-offensive.
In terms of the content preservation,
for both datasets our method also produces better results (the closer to 1 the better) when compared to \cite{shen17} .
A reason for these good results can be found by checking the examples presented in Table \ref{table:transf_examp}.
The use of the back transfer loss and the attention mechanism makes our model good at preserving the original sentence content while being precise at replacing offensive words by the non-offensive ones.
Also observe from Table \ref{table:transf_examp} that,
quite often, \citet{shen17}'s model changes many words in the original sentence, significantly modifying the content.

On the other hand, our model produces worse results in terms of perplexity values.
We believe this can be due to one type of mistake that is frequent among the transferred sentences and that is illustrated in Table \ref{table:mistakes}. The model uses the same non-offensive word (e.g. \emph{big}) to replace an offensive word (e.g. \emph{f***ing}) almost everywhere, which produces many unusual and unexpected sentences.

We have performed ablation experiments by removing some components of the proposed model. The results for the Twitter dataset are shown in Table \ref{table:ablation}.
We can see that attention and back-transfer loss play important roles in the model. In particular, when both of them are removed (last row in Table \ref{table:ablation}), although the classification accuracy improves, the perplexity and the content preservation drop significantly. This behavior happens due to the trade off that the decoder has to balance when transferring a sentence from a style to another.
The decoder must maintain a proper balance between transferring to the correct style and generating sentences of good quality. 
Each of these properties can easily be achieved on its own, e.g., 
copying the entire input sentence will give low perplexity and good content preservation but low accuracy, 
on the other hand, outputting a single keyword can give high accuracy but high perplexity and low content preservation.
While the classification loss guides the decoder to generate sentences that belong to the target style, 
the back transfer loss and the attention mechanism encourage the decoder to copy words from the input sentence.
When both back transfer loss and attention are removed, the model is encouraged to just meet the classification requirement in the transfer step.  
 
%, there is a trade off that the decoder has to balance between copying the input sentence (perfect perplexity/reconstruction) and modifying the sentence (which can lead to perfect classification).
%This happens because the generated sentences now mostly contain repetitions of the same few words, which can still make the classifier to believe that the sentences have correct sentiment but they become degenerate. This also highlights, in the task of unsupervised style transfer, the importance of maintaining a proper balance between the correct transferred style and the quality of the generated sentence. Each of this properties can easily be achieved on its own, e.g., copying an input sentence will give a good perplexity but low accuracy and poor content preservation, on the other hand, outputting a single keyword can give high accuracy but high perplexity. To have all of these properties in the generated sentence is a challenge and our proposed algorithm provides a viable step towards such a solution.

\begin{table}[!ht]
\begin{center}
\begin{tabular}{|l|rrr|}
\hline \bf System &  \bf Acc. & \bf CP  & \bf PPL \\ \hline
Full                   &  99.63 & \bf 0.947 & \bf 162.75 \\
\hline
No Attention            & 99.88 & 0.939 & 196.65 \\
No Back Transfer        & 97.08 & 0.938 & 257.93 \\
No Att \& Back Trans & \bf 100.0 & 0.876 & 751.56 \\
\hline
\end{tabular}
\end{center}
\caption{Ablation results for the Twitter dataset.}
\label{table:ablation}
\end{table}

It is important to note that current unsupervised text style transfer approaches can only handle well cases where the offensive language problem is lexical (such as the examples shown in Table \ref{table:transf_examp}), and just changing/removing few words can solve the problem.
The models experimented in this work will not be effective in cases of implicit bias where ordinarily inoffensive words are used offensively.

%% file: conclusions.tex
\section{Conclusions}
This work is a first step in the direction of a new promising approach for fighting abusive posts on social media.
Although we focus on offensive language,
we believe that further improvements on the proposed methods will allow us to cope with other types of abusive behaviors.
%In the type of posts analyzed, many times just the re